\documentclass[journal]{IEEEtran}
\IEEEoverridecommandlockouts

\usepackage{cite}
\usepackage{graphics} 
\usepackage{epsfig} 
\usepackage{mathptmx} 
\usepackage{times} 
\usepackage{amsmath} 
\usepackage{amssymb}  
\usepackage{multirow}
\usepackage{booktabs}
\usepackage{fourier-orns}
\usepackage{setspace}
\usepackage{graphicx}
\usepackage[justification=centering]{caption}
\usepackage{float}
\usepackage{verbatim}
\usepackage{mathrsfs}

\begin{document}
	
\title{STGIN: A Spatial Temporal Graph-Informer Network for Long Sequence Traffic Speed Forecasting}

\author{Ruikang~Luo,~\IEEEmembership{Student Member,~IEEE,}
	    Yaofeng~Song,
	    Liping~Huang,~\IEEEmembership{Member,~IEEE,}
   	    Yicheng~Zhang,~\IEEEmembership{Member,~IEEE,}
        and~Rong~Su,~\IEEEmembership{Senior Member,~IEEE}
        
\thanks{Ruikang Luo is affiliated with Continental-NTU Corporate Lab, Nanyang Technological University, 50 Nanyang Avenue, 639798, Singapore Email: ruikang001@e.ntu.edu.sg}
\thanks{Yaofeng Song is affiliated with School of Electrical and Electronic Engineering, Nanyang Technological University, 639798, Singapore Email: song0223@e.ntu.edu.sg@e.ntu.edu.sg}
\thanks{Liping Huang is affiliated with School of Electrical and Electronic Engineering, Nanyang Technological University, 639798, Singapore Email: liping.huang@ntu.edu.sg}
\thanks{Yicheng Zhang is affiliated with Institute for Infocomm Research (I2R), Agency for Science, Technology and Research (ASTAR), 138632, Singapore Email: zhang$\_$yicheng@i2r.a-star.edu.sg}%
\thanks{Rong Su is affiliated with Division of Control and Instrumentation, School of Electrical and Electronic Engineering, Nanyang Technological University, 50 Nanyang Avenue, Singapore 639798. Email: rsu@ntu.edu.sg}
}

\markboth{IEEE TRANSACTIONS ON INTELLIGENT TRANSPORTATION SYSTEMS}%
{Shell \MakeLowercase{\textit{et al.}}: Bare Demo of IEEEtran.cls for IEEE Journals}

\maketitle

\begin{abstract}
Accurate long series forecasting of traffic information is critical for the development of intelligent traffic systems. We may benefit from the rapid growth of neural network analysis technology to better understand the underlying functioning patterns of traffic networks as a result of this progress. Due to the fact that traffic data and facility utilization circumstances are sequentially dependent on past and present situations, several related neural network techniques based on temporal dependency extraction models have been developed to solve the problem. The complicated topological road structure, on the other hand, amplifies the effect of spatial interdependence, which cannot be captured by pure temporal extraction approaches. Additionally, the typical Deep Recurrent Neural Network (RNN) topology has a constraint on global information extraction, which is required for comprehensive long-term prediction. This study proposes a new spatial-temporal neural network architecture, called Spatial-Temporal Graph-Informer (STGIN), to handle the long-term traffic parameters forecasting issue by merging the Informer and Graph Attention Network (GAT) layers for spatial and temporal relationships extraction. The attention mechanism potentially guarantees long-term prediction performance without significant information loss from distant inputs. On two real-world traffic datasets with varying horizons, experimental findings validate the long sequence prediction abilities, and further interpretation is provided.

\end{abstract}


\IEEEpeerreviewmaketitle

\section{Introduction}
\IEEEPARstart{T}{he} intelligent traffic system is a significant part of the modern city. Traffic congestion alleviation is one of the challenges faced by traffic management, traffic planning and traffic control\cite{xiong2017optimal}. Traffic information usually includes traffic flow, link speed, traffic density, vehicle type distribution\cite{luo2022dense} and facility usage conditions\cite{luo2020traffic}\cite{song2022sumo}. Long-term accurate traffic forecasting is critical for traffic scheduling and road condition assessments\cite{huang2022incremental}, which are two major components of the intelligent traffic system. As the increment of data sensors deployment\cite{luo2019mission}, huge amount of real-time data could be collected to help the traffic planner\cite{meng2017city}, and the requirements for data processing technology are also getting higher\cite{theofilatos2017incorporating}.

However, predicting traffic has gotten more difficult as a result of its complicated spatial-temporal connections\cite{luo2022ast}. Firstly, the shifting patterns in driving behavior are unknown in the temporal dimension. The traffic parameter varies on a periodic basis and is relatively easy to describe in the long run, as illustrated in Fig.1. Traditional techniques, on the other hand, rely heavily on stationary circumstances. Sudden events\cite{siami2018comparison}, such as accidents and emergencies, can make the system difficult to quickly response such changes. Secondly, the complete spatial relationships between roads and cars contribute significantly to the future state. For instance, if a traffic accident occurs on the connection, congestion will occur. Upstream vehicles may alter their path, and nearby connections may experience more diversion in subsequent stages. The hidden pattern cannot be extracted purely by temporal dependencies\cite{li2017diffusion}.

\begin{figure}[!htb]
	\centering
	\includegraphics[width=1\linewidth]{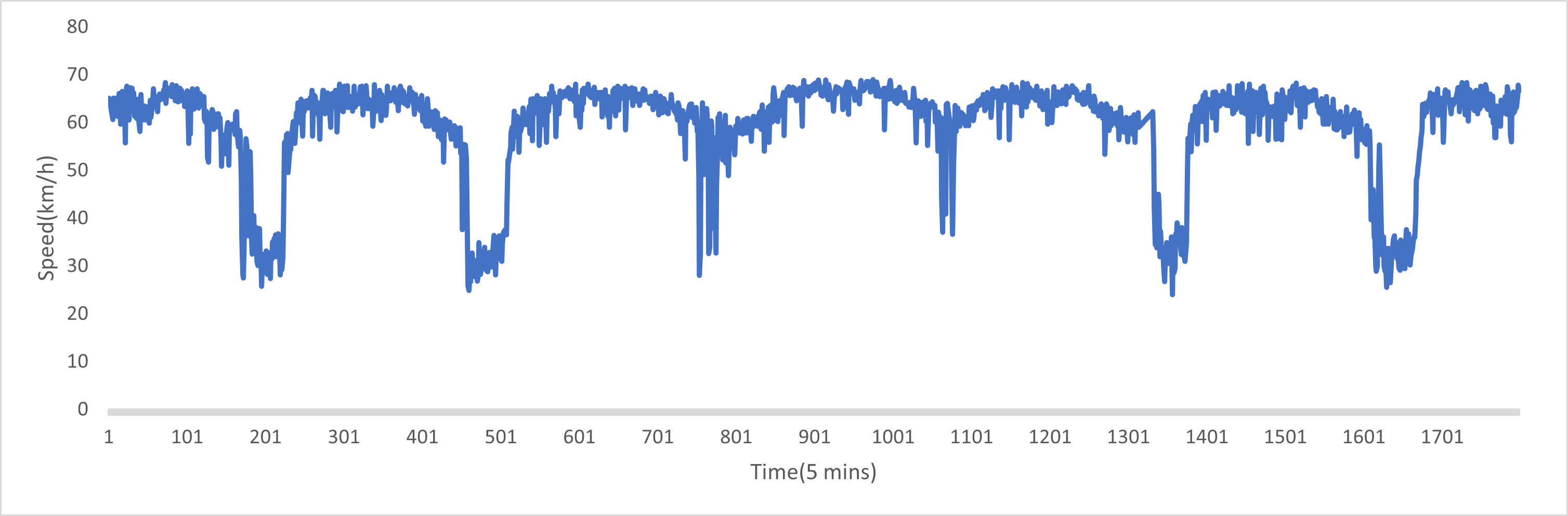}
	\caption{Traffic periodicity}
	\label{fig00}
\end{figure}

Numerous existing techniques for addressing these issues may be classified into two categories: model-driven methods, which employ dynamic mathematical models to replicate driving behavior, and data-driven methods, such as Auto-Regressive Integrated Moving Average (ARIMA) model\cite{ahmed1979analysis}\cite{hamed1995short}, Kalman filtering\cite{okutani1984dynamic}, Gated Recurrent Unit (GRU) network and Graph Neural Network (GNN). Among them, RNN models have been extensively applied to train learning models from sequential data\cite{lv2018lc}. However, RNN models treat different roads data independently without considering spatial connectivity\cite{zhang2018combining}. To learn the spatial features from the given network, some studies propose Convolutional Neural Network (CNN)\cite{zhang2017deep} for spatial dependencies modeling, however, the performance of CNN for non-Euclidean data is not satisfying\cite{cao2017interactive}. Thus, GNN-based methods are proposed recently to understand the spatial-temporal information by taking time-series road data and connections between roads. These models usually combine RNN and GNN to adapt the complicated forecasting task.

GNN-based spatial-temporal models have been proven to have better performance on some forecasting problems than pure RNN or GNN models, there are still some limitations. RNNs extract the temporal dependencies by traversing the information over a long recurrent chain bidirectionally\cite{connor1994recurrent}, which is difficult to obtain information from a distant data and cannot avoid the vanishing gradient issue\cite{lv2014traffic}. Moreover, traffic parameters do not just sequentially depend on the past and current data. The periodicity can be impacted continuously by the emergency. Thus, it is important to understand the global dependencies over time.

To address these issues, this study proposes a novel spatial-temporal Transformer-based model for forecasting long series traffic information. It combines a GAT layer with an Informer layer, which was recently introduced for long sequence forecasting via an attention mechanism, to more effectively extract complicated topological road structure and global temporal relationships. There are three main contributions as follows:

\begin{itemize}
\item The novel STGIN model containing the GAT layer and Informer layer is proposed to better capture adjacent roads information and global temporal dependencies for long sequence traffic forecasting. Based on the existing knowledge, the proposed method has the ability to attend more than sequential connection and achieve global understanding of the hidden traffic pattern.
\item The study performs extensive analysis regarding long sequence predictive ability over different horizons. The forecasting results of the STGIN show a steady performance and much better performance than other baselines when the target sequence is longer.
\item Two real-world traffic speed datasets and one traffic facility usage condition dataset are deployed for performance evaluation. The results show over 85$\%$ predictive accuracy and around 10$\%$ accuracy improvement than baseline algorithms. The feasibility has been verified.
\end{itemize}

In the next several sections, the content is organized as follows. In Section Two, recent research on traffic forecasting are introduced for comparison. In Section Three, the proposed STGIN framework is described in detail. In Section Four, experiments settings and the predictive performance is evaluated and visualized under various metrics with other baseline models. In Section Five, we give the conclusion and future plan.

\section{Related Work}
Existing traffic prediction methods can be summarized as the model-driven and the data-driven methods\cite{zhao2020attention}. For the model-driven methods, they use stationary mathematics model, such as cell transmission model\cite{wei2013total}, queuing theory model\cite{xu2014analysis}, microscopic fundamental diagram model\cite{xu2013impacts}, to express the instantaneous traffic relationships. However, a wealth of prior knowledge is required to predict the traffic condition precisely and these methods usually cannot take unexpected events and external factors into account. Data-driven approaches\cite{silver2016mastering}, which are our concerns, can be further divided into classic machine learning models and neural network based models. Next, four related methods: autoregressive model, RNN model, GNN model and Transformer-based model, will be discussed in detail.

The representative methods of traditional autoregressive models contain VAR, SVR\cite{yao2006research} and ARIMA models\cite{luo2019traffic}. VAR and SVR require stationary sequential data, and ARIMA is commonly used in the early time to process non-stationary time-series data\cite{sun2004interval}, but does not perform well on long-term prediction\cite{dudek2016pattern}.

Since deep learning models were firstly applied to forecast traffic in 2014\cite{huang2014deep}, RNN and its variant models such as LSTM networks\cite{hochreiter1997long} and GRU networks\cite{cho2014learning} were found popular to learn comprehensive dynamic schemes from sequential traffic data and other scenarios. In some research, authors compared different RNN models, finding GRU performed better than LSTM in the detailed traffic forecasting problem\cite{fu2016using}. Further, to solve the problem of capturing forward temporal dependencies, authors proposed the stacked bidirectional and unidirectional LSTM structure\cite{cui2018deep}. But the inherent limits of losing distant data information and spatial dependencies still cannot be solved.

To extract spatial correlations in traffic prediction, CNNs and GNNs have been widely used. Though CNNs show good performance in capturing topological dependencies in images, videos and other Euclidean data\cite{huang2017study}\cite{yang2020domain}, road network is in the form of non-Euclidean data, which cannot be processed well by CNNs\cite{defferrard2016convolutional}. Some research converted network-wide traffic matrices into images for CNNs application\cite{ke2017short}, however the overall effect was inefficient. GNNs directly process the road network as a graph with edges and nodes to extract spatial dependencies\cite{wu2020comprehensive}\cite{jiang2021graph}, which achieve satisfying results. As the recent development, it can be categorized into Diffusion Graph Convolution (DGC)\cite{atwood2016diffusion}, Graph Convolutional Network (GCN)\cite{kipf2016semi} and GAT\cite{velivckovic2017graph}. Among them, GAT provides a good insight to capture spatial information and commonly combined with different temporal extraction layers to form spatial-temporal models under various scenarios. For example,  $ST-MetaNet^+$\cite{pan2020spatio} combines GAT and RNN to predict traffic speed information with fixed relationship between road nodes. DGC models the spatial dependencies as a diffusion process to further adapt to real road conditions. To learn the real road network connections, the attention mechanism is used in GAT to understand the real road network connections instead of predefining relative weights between two nodes. But it still has the limitation that the proposed temporal attention mechanism cannot fully consider the complex realities of the road.

Recently, Transformer with self-attention mechanism has been widely concerned as a new deep learning architecture for sequence modeling and verified more effective than RNNs in some scenarios. It applies attention mechanism to score connections between each input and output positions pair without information loss\cite{liu2018generating}\cite{parmar2018image}. Several studies have proposed to utilize Transformer for time-series forecasting problem\cite{wang2020traffic}\cite{cai2020traffic}. Specifically, in this work\cite{li2019enhancing}, authors added a convolutional self-attention layer to enhance performance on local contexts and proved effectiveness in long-term dependencies extraction. However, it ignored spatial dependencies of the road structure and training data requirement is huge. Authors have also proposed the idea of Transformer to deal with taxi ride-hailing forecasting problems by using sparse linear layers to better capture spatial dependencies\cite{li2019forecaster}. Lately, in this work\cite{zhou2021informer}, authors proposed a novel Transformer-based architecture, called Informer, to address long sequence time-series forecasting problem by using ProbSparse Self-attention. The model has been verified on diverse datasets and achieved better performance on prediction accuracy and time complexity than other Transformer variants. But similarly, the spatial dependencies extraction was ignored and network-wide traffic forecasting cases have not been verified. 

Motivated by these studies, a novel spatial-temporal Transformer-based network is proposed in this paper to explicitly model the spatial correlations and the periodicity of the traffic network.

\section{Methodology}
\subsection{Problem Analysis}
The purpose of this research is to predict the traffic information in the long-term based on historical records and topological road network. Traffic information on roads is a general concept and could be link volume, vehicle density, link speed, and facility usage status. The road network structure can be denoted as a directed graph $G=\{O, R, A\}$. Each location recording the data is represented by a node combining with features, and $O$ is the nodes set; $R$ is edges set connecting nodes in the road graph; the adjacency matrix $A \in R^{N \times N}$, which represents paired nodes connectivity, where $N$ is the number of nodes. Instead of assigning elements of 0 and 1 to indicate connections between two nodes, the adjacency matrix elements are calculated by the real road distances between nodes using Gaussian kernel weighting function\cite{zhao2019t}:
\begin{equation}
A_{ab} = 
\left\{
\begin{array}{lr}
\text{exp}(-\frac{len(v_a, v_b)^2}{\sigma^2}),\ len(v_a, v_b) \leq \kappa &\\
0,\ \text{otherwise} &
\end{array}
\right.
\end{equation}
where $len(v_a, v_b)$ is the actual road length from node $v_a$ to node $v_b$, instead of the Euclidean distance. $\sigma$ is the standard deviation of road length; $\kappa$ is the preset threshold to filter negligible elements.

The feature matrix is used to represent historical attributes of $N$ nodes at time $t$ on the road graph, which could be represented by $\textbf{X}_t=\{\vec{x}_{1,t}, \vec{x}_{2,t},...,\vec{x}_{N,t} \}$, where $\textbf{X}_t\in R^{N \times \mathscr{F}}$, $\vec{x}_a \in R^{\mathscr{F}}$ is $a-th$ input node features and $\mathscr{F}$ is the feature dimension of the historical data. Now the problem can be treated as: given the total known $E$ historical and current observation sequence $\left\{ \textbf{X}_{t-E+1}, ..., \textbf{X}_t \right\}$ as input and the traffic network graph $G$ constructed, we need to find a mapping function $f$ to predict the future $F$ time steps traffic information $\left\{\hat{\textbf{Y}}_{t+1}, ..., \hat{\textbf{Y}}_{t+F}\right\}$, where $\hat{\textbf{Y}}_{t+F}$ means the traffic information estimated at time $t+F$ in the future as equation 2: 
\begin{equation}
\left\{\hat{\textbf{Y}}_{t+1}, ..., \hat{\textbf{Y}}_{t+F}\right\} = f (G; \left\{\textbf{X}_{t-E+1}, ..., \textbf{X}_t\right\})
\label{learning_function}
\end{equation}

\subsection{Feature Convolution Aggregator}
Before introducing STGIN framework, appropriate data processing for the traffic data and external factors is essential for information aggregation. Our design is to use multiple convolutional operators treating every input link separately to aggregate the traffic speed profile together with all external factors. Correspondingly, there are multiple GAT networks that intakes the aggregated data generated by convolutional networks. This data linear transformation structure is called Feature Convolution Aggregator (FCA).

\subsection{GAT-Spatial Dependence Modeling}
CNN shows good performance when dealing with Euclidean problem, such as image recognition, however, the real urban traffic network is a kind of typical non-Euclidean data in the form of a graphic structure\cite{luo2021deep}. GAT remains the attention mechanism to indicate the importance between every two nodes. The architecture of GAT is shown in Fig. 2.

\begin{figure}[!htb]
	\centering
	\includegraphics[width=1\linewidth]{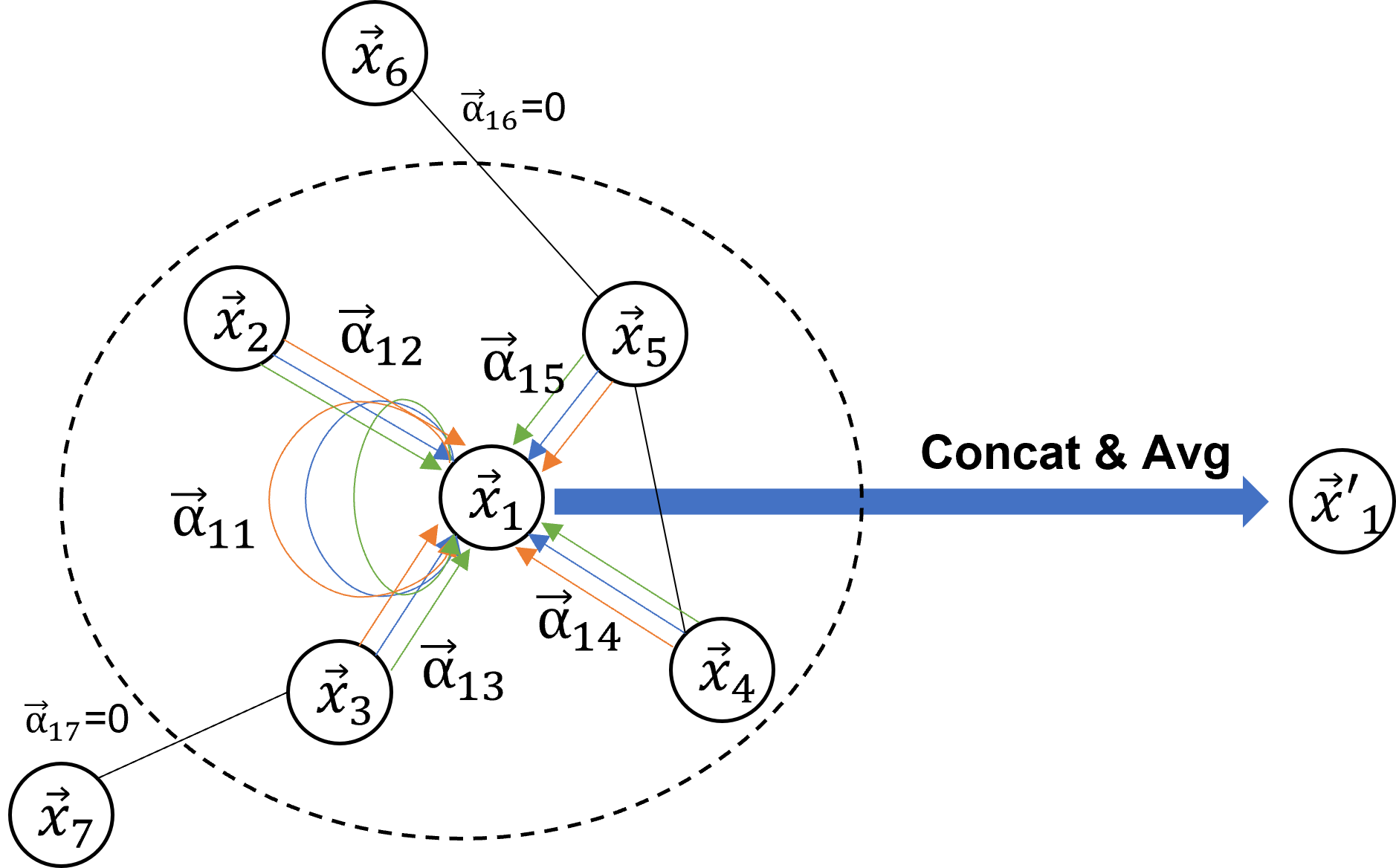}
	\caption{The multi-head GAT layer illustration}
	\label{fig01}
\end{figure}

In general, network $G$ contains relationships among roads. The neighboring roads are likely to affect each other and share the similar attributes. For example, the traffic volume on the downstream link in the next time step is highly dependent on the current traffic volume on the upstream link. Hence, to capture the spatial dependence, GAT model is proposed to transform and propagate information as the first layer.

GAT layer utilizes FCA to reflect the input features into high-dimensional features over nodes of the graph to captures spatial information. The linear transformation can be expressed by a weighted matrix, $\textbf{W} \in R^{\mathscr{F}^{\prime} \times \mathscr{F}}$, which is applied over nodes. The attention coefficients can be computed by\cite{velickovic2017graph}:
\begin{equation}
e_{ab}= att(\textbf{W} \vec{x}_a,\textbf{W} \vec{x}_b)
\end{equation}
where $\mathscr{F}^{\prime}$ is the number of output features. As stated, the coefficients indicate the correlation between every two nodes.

However, in the real traffic network, not every two links are connected. Thus, the adjacency matrix is used to select the topological neighborhood in the graph for attention coefficient computation. In this study, suitable normalization (Softmax) and activation function (LeakyReLU) are applied to generate the full attention mechanism:
\begin{equation}
\alpha_{ab}=\frac{{\rm exp}({\rm LeakyReLU}(\vec{\textbf{att}}^T[\textbf{W}\vec{x}_a \Vert \textbf{W} \vec{x}_b]))}{\sum_{c\in\mathscr{N}_a} {\rm exp}({\rm LeakyReLU}(\vec{\textbf{att}}^T[\textbf{W}\vec{x}_a \Vert \textbf{W} \vec{x}_c]))}
\end{equation}
where $\mathscr{N}_a$ is the topological neighborhood of node $a$ in the graph, $\cdot^T$ represents transposition operation and $\Vert$ is the concatenation operation.

Further, the computed coefficients are used to find the final output features for every node. Similar to multi-head attention operation in Transformer\cite{vaswani2017attention}, the averaging result is served as the final output:
\begin{equation}
\vec{x}_a^\prime = \sigma(\frac{1}{C}\sum_{c=1}^{C}\sum_{j\in\mathscr{N}_a}\alpha_{ab}^c\textbf{W}^c\vec{x}_b)
\end{equation}

\subsection{Informer-Temporal Dependence Modeling}
\subsubsection{Canonical Transformer and Self-attention Mechanism}
Acquiring global temporal dependency is a critical step in forecasting traffic information. Currently, neural network-based methods are frequently utilized to analyze and predict sequence data. However, the deficiencies discussed in the preceding section limit the long-term forecasting performance. Thus, one model based on transformers, Informer, is proposed to collect global temporal information\cite{zhou2021informer}. Informer, like the canonical transformer model, employs an encoder-decoder architecture.

\begin{figure}[!htb]
	\centering
	\includegraphics[width=0.95\linewidth]{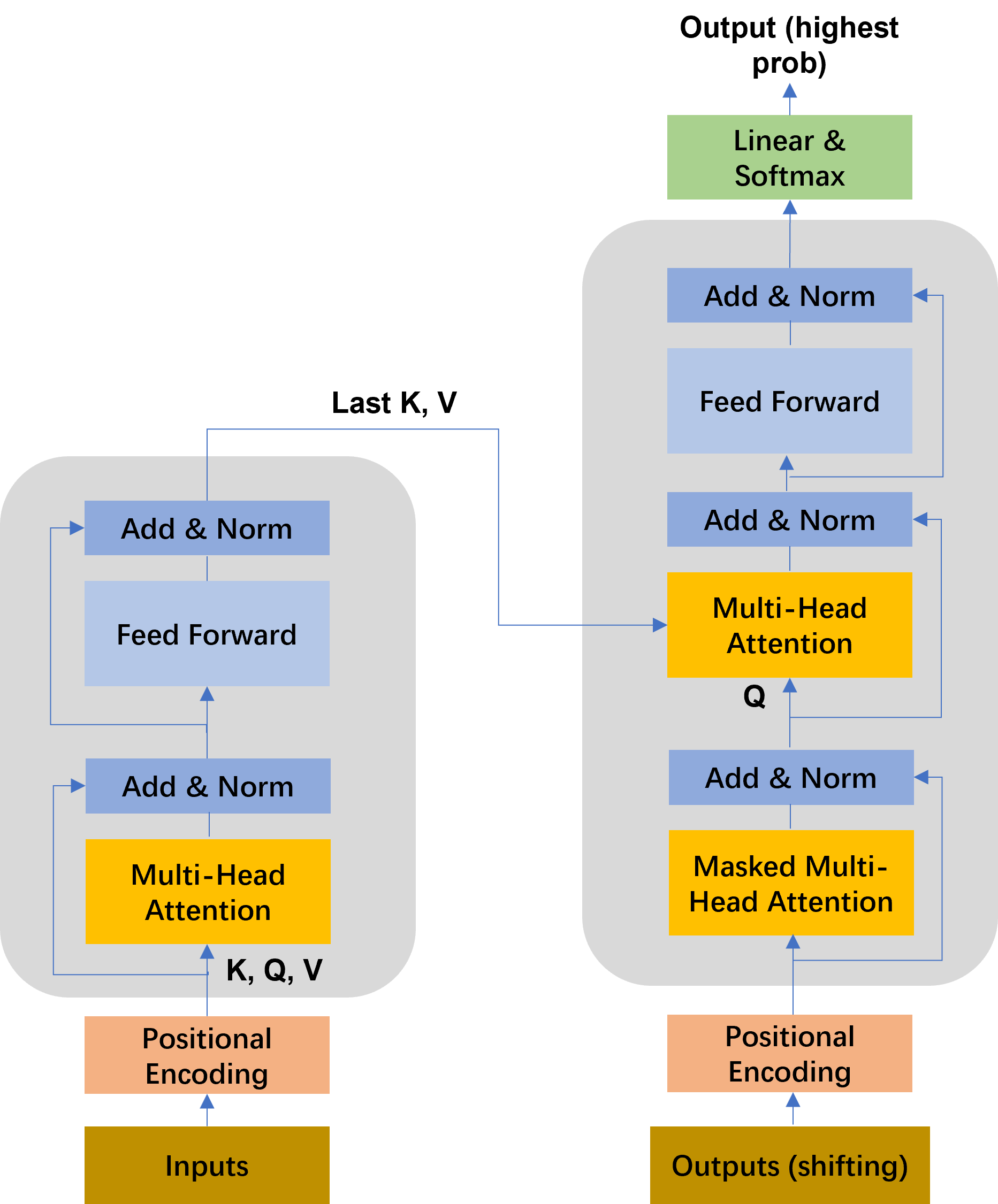}
	\caption{Transformer architecture}
	\label{fig02}
\end{figure}

The general Encoder-Decoder type architecture of transformer is shown as Fig. 3.  In transformer\cite{vaswani2017attention}, each element is paired to all other input sequence elements. The purpose of this exercise is to determine the distribution of attention and the most closely connected pairings. The self-attention mechanism is characterized as executing a scaled dot-product on all nodes' queries, keys, and values and calculating attention scores for each location using a softmax function on aggregated data. The attention is calculated as follows:
\begin{equation}
A(Q, K, V)=Softmax(\frac{QK^T}{\sqrt{d}})V
\label{self-attention}
\end{equation}
where $Q$ $\in$ $R^{l_Q \times d_{in}}$, $K$ $\in$ $R^{l_K \times d_{in}}$, $V$ $\in$ $R^{l_V \times d_{in}}$ and $d_{in}$ is the dimension of input sequence.

Let $q_i$, $k_i$, $v_i$ represent the $i-th$ row in $Q$, $K$, $V$, then specificly, the $i-th$ query's attention is defined\cite{tsai2019transformer}:
\begin{equation}
A(q_i, K, V)=\sum_{j}\frac{k(q_i,k_j)}{\sum_{l}k(q_i,k_l)}v_j=\sum_{j}p(k_j|q_i)v_j=E_{p(k_j|q_i)}[v_j]
\label{queryattention_prob}
\end{equation}

Even though the dot-product operations are executed for every elements, in most real-world scenarios, only several significant dot-product pairs determine the dominated attention. A suitable selective mechanism for dot-product pairs that can distinguish them has the potential to improve the prediction capacity.

\subsubsection{ProbSparse Self-attention}
From above derivations, the known $i-th$ query's attention towards other keys is notated as a probability $p(k_j|q_i)$. The dominant dot-product pairs trend to show the separation between the uniform distribution $q(k_j|q_i)=\frac{1}{L_k}$ and the corresponding query's attention probability distribution. Kullback-Leibler divergence\cite{van2014renyi} is widely utilized to evaluate the correlation between distributions p and q as:
\begin{equation}
KL(q||p) = {\rm ln}\sum_{j=1}^{l_K}e^\frac{{q_{i}k_{l}^{T}}}{\sqrt{d_{in}}}-\frac{1}{l_K}\sum_{j=1}^{l_K}\frac{q_{i}k_{j}^{T}}{\sqrt{d_{in}}}-{\rm ln}l_K
\label{Kullback-Leibler}
\end{equation}

Based on Eq.(), the i-th query's sparsity measurement is defined as:
\begin{equation}
M(q_i, K) = {\rm ln}\sum_{j=1}^{l_K}e^\frac{{q_{i}k_{l}^{T}}}{\sqrt{d_{in}}}-\frac{1}{l_K}\sum_{j=1}^{l_K}\frac{q_{i}k_{j}^{T}}{\sqrt{d_{in}}}
\label{SparsityMeasurement}
\end{equation}

Further, since we have the bounds for each query in the keys set K:
\begin{equation}
{\rm ln}l_k \leq M(q_i, K) \leq max_{j}{\frac{q_{i}k_{j}^{T}}{\sqrt{d_{in}}}}-\frac{1}{l_K}\sum_{j=1}^{l_K}\frac{q_{i}k_{j}^{T}}{\sqrt{d_{in}}}+lnl_K
\label{bound}
\end{equation}

we have the approximate measurement as:
\begin{equation}
\bar{M}(q_i, K)=max_{j}{\frac{q_{i}k_{j}^{T}}{\sqrt{d_{in}}}}-\frac{1}{l_K}\sum_{j=1}^{l_K}\frac{q_{i}k_{j}^{T}}{\sqrt{d_{in}}}
\label{approximation}
\end{equation}

If $\bar{M}(q_i, K)$ is larger, the $i-th$ query attention probability is more diverse. And it is highly potential to have the dominate dot-product pairs.

Using this measurement, ProbSparse Self-attention is applied to let each key only paired with the dominant queries:
\begin{equation}
A(Q, K, V)=Softmax(\frac{\bar{Q}K^T}{\sqrt{d_{in}}})V
\label{ProbSparse Self-attention}
\end{equation}

where $\bar{Q}$ and $Q$ has the same size, however, $\bar{Q}$ only has the Top-u pairs under the sparsity measurement $\bar{M}(q_i, K)$ and other pairs filled with zero. Till this stage, the ProbSparse Self-attention has been constructed with more sensitive ability to potential dominant dot-product pairs.

\subsubsection{Self-attention Distilling}
Self-attention distilling operation is designed to extract dominating attention and construct a focused feature map. The distilling operation from m-th layer to (m+1)-th layer as:
\begin{equation}
X_{m+1}^{t}=MaxPool(ELU(Conv1d([X_{m}^{t}]_{AB})))
\label{Distilling}
\end{equation}

where Conv1d() performs an 1-D convolution on time dimension and $[]_{AB}$ represents the Multi-head ProbSparse self-attention operations. The max-pooling layer helps to reduce $X^t$ into its half slice in the next layer. Halving replicas can be stacked to improve the robustness and one self-attention layer is reduced at a time to ensure the aligned output dimension of the concatenated feature map, which is the hidden representative of encoder.

\begin{figure*}[!htb]
	\centering
	\includegraphics[width=1\linewidth]{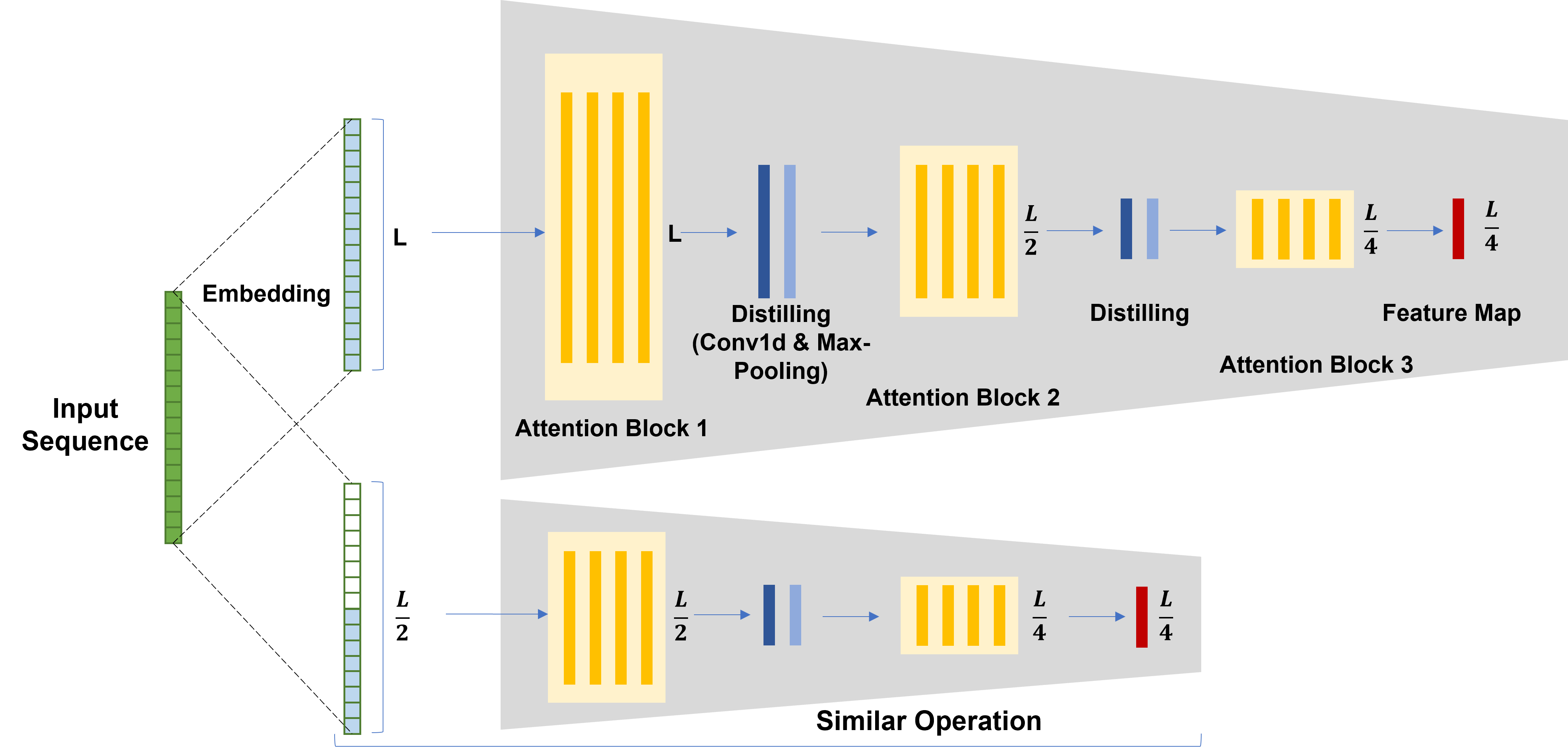}
	\caption{Encoder structure of Informer (Self-attention Distilling included)}
	\label{fig04}
\end{figure*}

\subsubsection{One-Forward Inference}
For the decoder of Informer, the similar structure with classical Transformer is applied, which contains two multi-head attention layers. The first layer will take the concatenated vector of start token, $X_{token}^t$, and placeholder, $X_{0}^t$ as the input, $X_{de}^t$. The masked operation is to avoid auto-regressive using the future stpes. And the second layer will take the feature map and outputs from the first layer as its inputs. In the end, a fully connected layer helps matching final sequence outputs.

In particular, a $L_{token}$ long input sequence ahead the target predictive window is chosen to as $X_{token}^t$. The length of $X_{0}^t$ is exact the length of target sequence time stamps. The decoder proposed in Informer can execute prediction by one forward inference, which further improves computational efficiency than classical Transformer model.

The completed architecture of Informer is as shown in Fig.5:
\begin{figure}[H]
	\centering
	\includegraphics[width=1\linewidth]{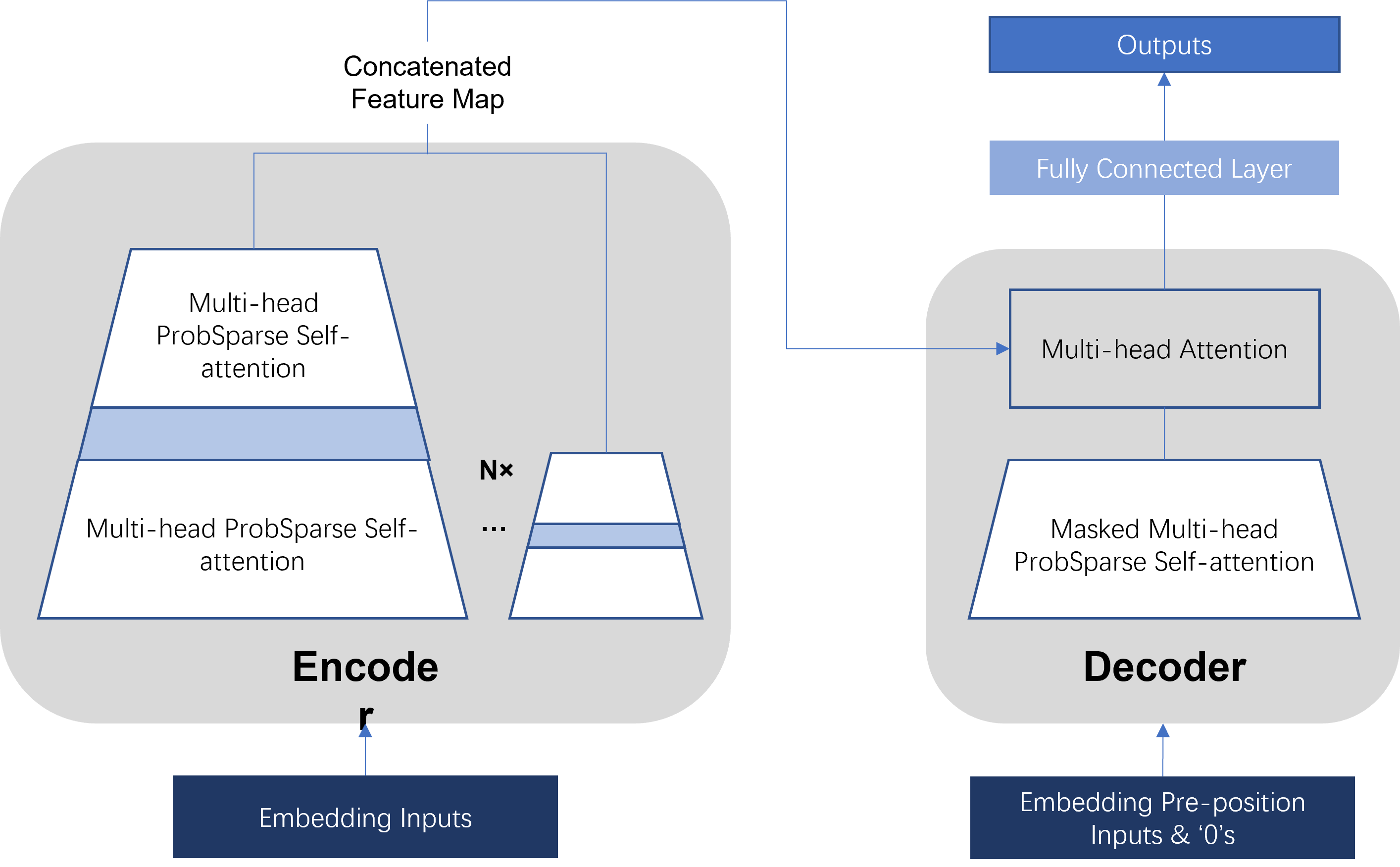}
	\caption{Informer architecture}
	\label{fig03}
\end{figure}

\subsection{STGIN-Overview Model}

\begin{figure*}[!htb]
	\centering
	\includegraphics[width=1\linewidth]{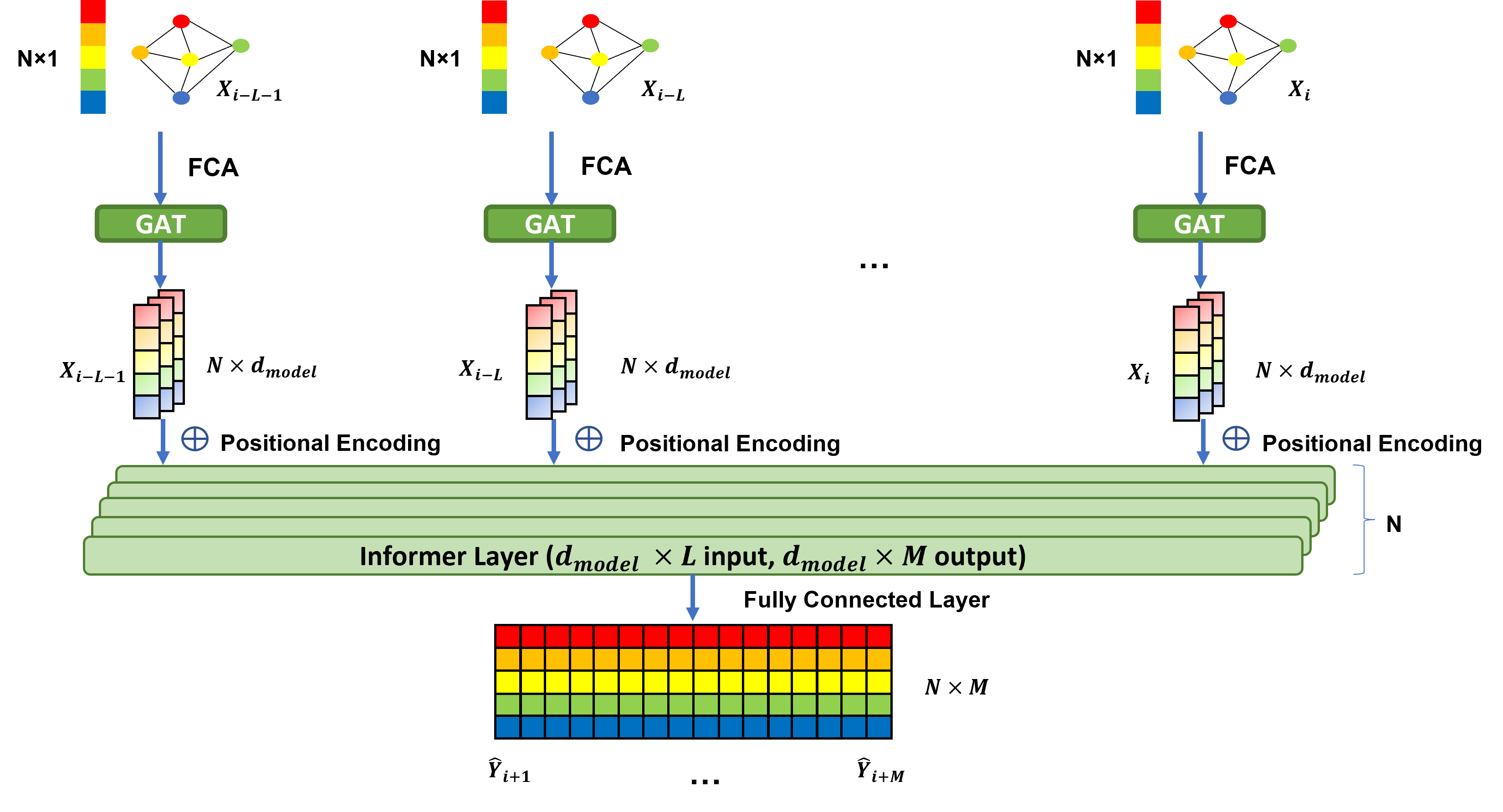}
	\caption{STGIN overview architecture}
	\label{fig05}
\end{figure*}

The overview architecture of STGIN model is shown in Fig.6. STGIN contains two parts: 1) the spatial GAT layers, which capture the spatial relations among traffic road networks; 2) Informer layers, which capture the long sequence temporal dependence. GAT layers are utilized to propagate spatial information through the traffic road networks. Thus, the outputs of GAT layers for T historical instances, which each contains N-node slices information, are fed as the inputs of Informer layers. Similarly, there are N Informer units used to process the historical sequential inputs of N-node structure. Finally, the outputs of N Informer units form the final prediction result in the shape of N-node by $T^{\prime}$ future time steps. In the diagram, each hidden representation for one past time stamps is expressed by a three-slice combination, actually the number is exactly the node number. d is the embedding dimension, which can be set freely.

\section{Experiments}
We conduct experiments and visualization to further demonstrate the predictive performance of STGIN model. The selected datasets are introduced first and then evaluation metrics are stated. Next, experiment settings and baseline models are described. In the end, the experiment results and detailed discussion are given to deeply understand the framework.

\subsection{Data Description}
There are two real-world vehicle speed datasets utilized for our experiments evaluation: SZ-taxi dataset and Los-loop dataset.
\begin{itemize}
	\item SZ-taxi: This dataset was collected from Luohu District, Shenzhen for taxi vehicles. 156 major roads are selected and taxi speed values are recorded from Jan. 1 to Jan. 31, 2015. The records are aggregated every 5 minutes. Apart from the feature matrix, the dataset also contains the adjacency matrix, which shows the spatial dependence between 156 roads.
	
	\item Los-loop: Los-loop dataset was collected by loop detectors at the highway of Los Angeles County. Among them, 207 sensors are selected and traffic speed values are recorded from Mar. 1 to Mar. 7, 2012. Records are aggregated every 5 minutes. The corresponding adjacency matrix, which is derived by the road length between sensors, is affiliated. Proper technics, such as linear interpolation, are utilized to fill missing data.	
\end{itemize}

\begin{figure}[!htb]
	\centering
	\includegraphics[width=0.9\linewidth]{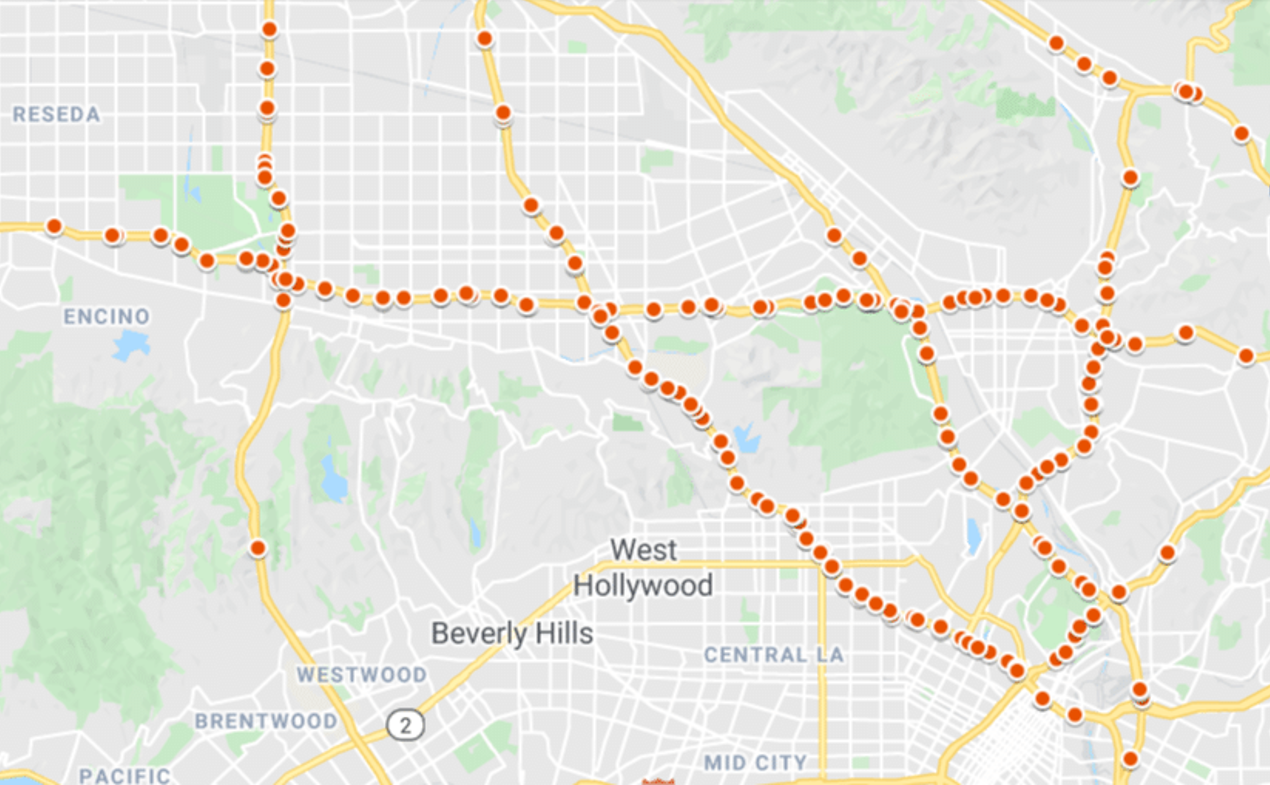}
	\caption{Sensors distribution on the highway of Los Angeles County}
	\label{fig52}
\end{figure}

\begin{figure}[!htb]
	\centering
	\includegraphics[width=0.9\linewidth]{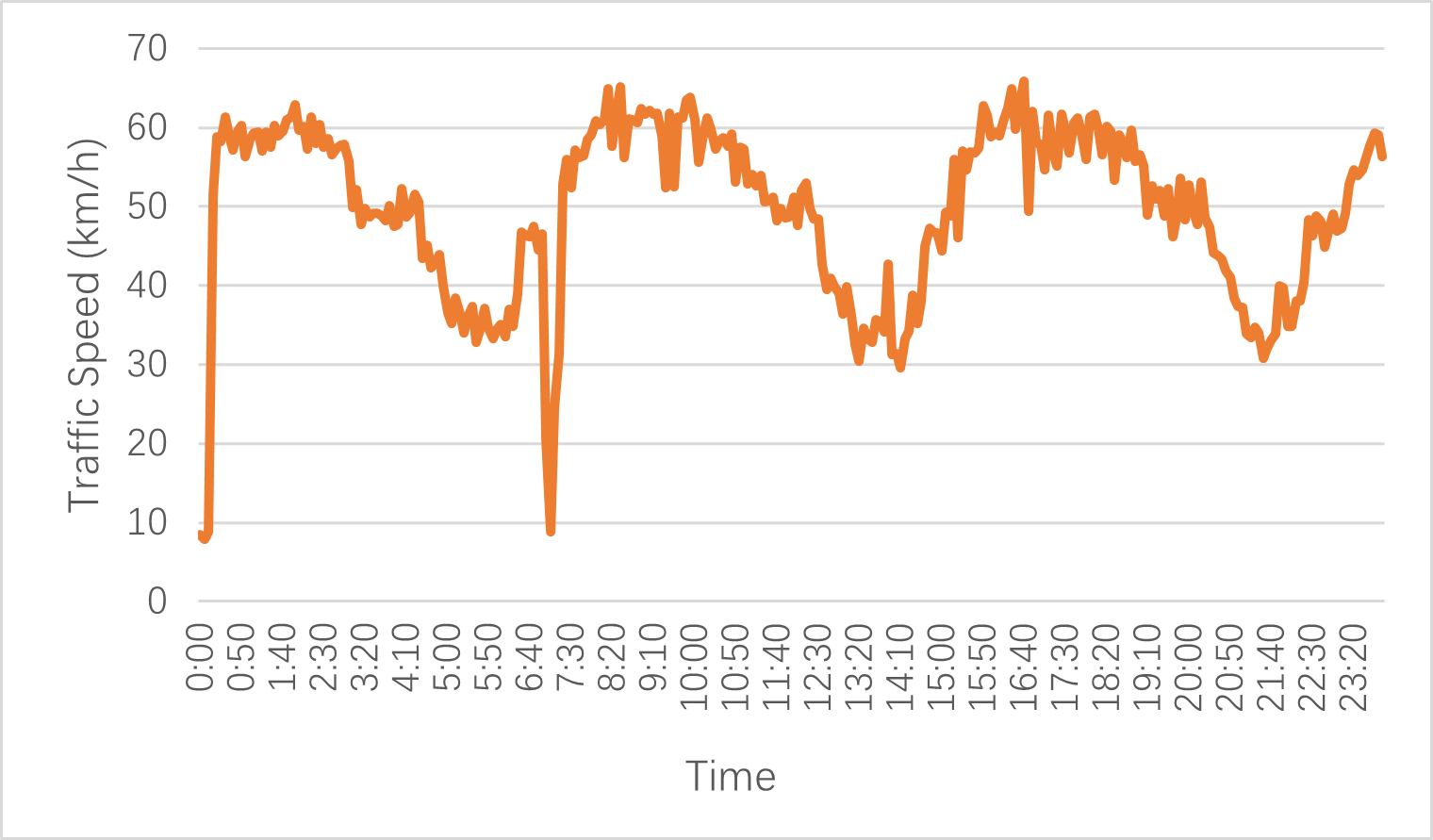}
	\caption{Traffic speed on Jan. 1st, Shenzhen}
	\label{fig09}
\end{figure}

\begin{figure}[!htb]
    \centering
    \includegraphics[width=0.9\linewidth]{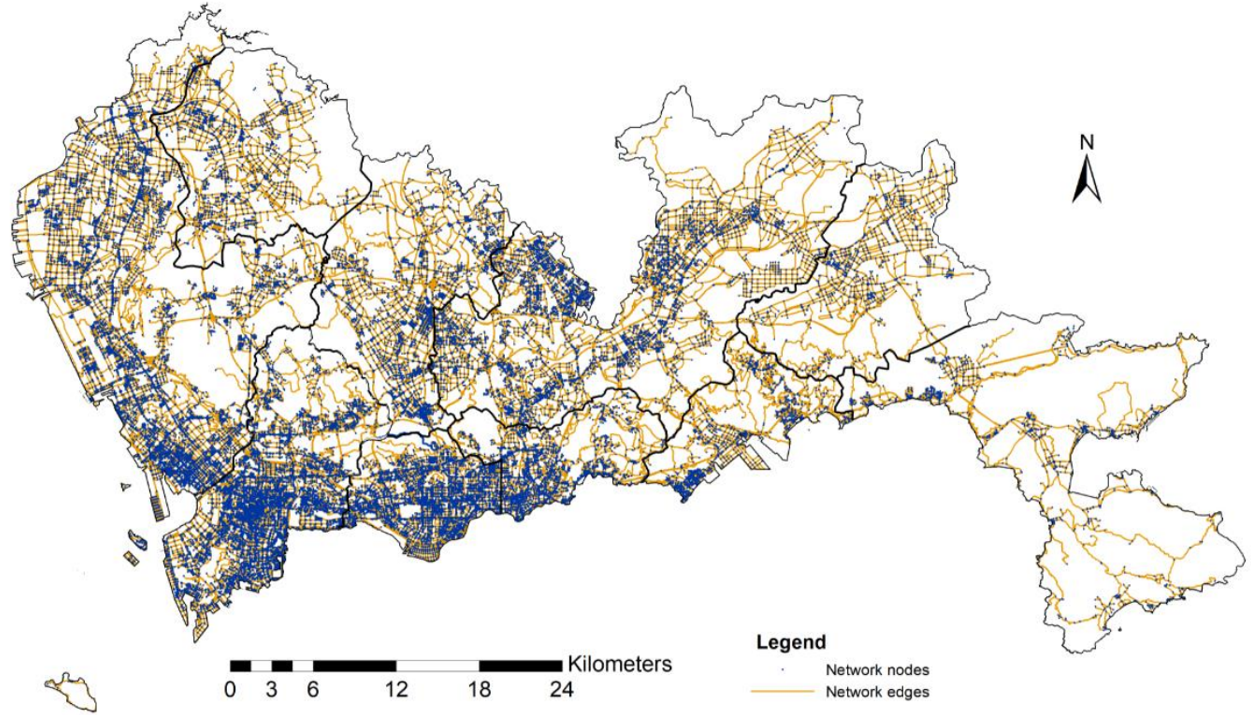}
    \caption{Digital representation of road network in Shenzhen}
	\label{fig53}
\end{figure}

\begin{figure}[!htb]
	\centering
	\includegraphics[width=0.9\linewidth]{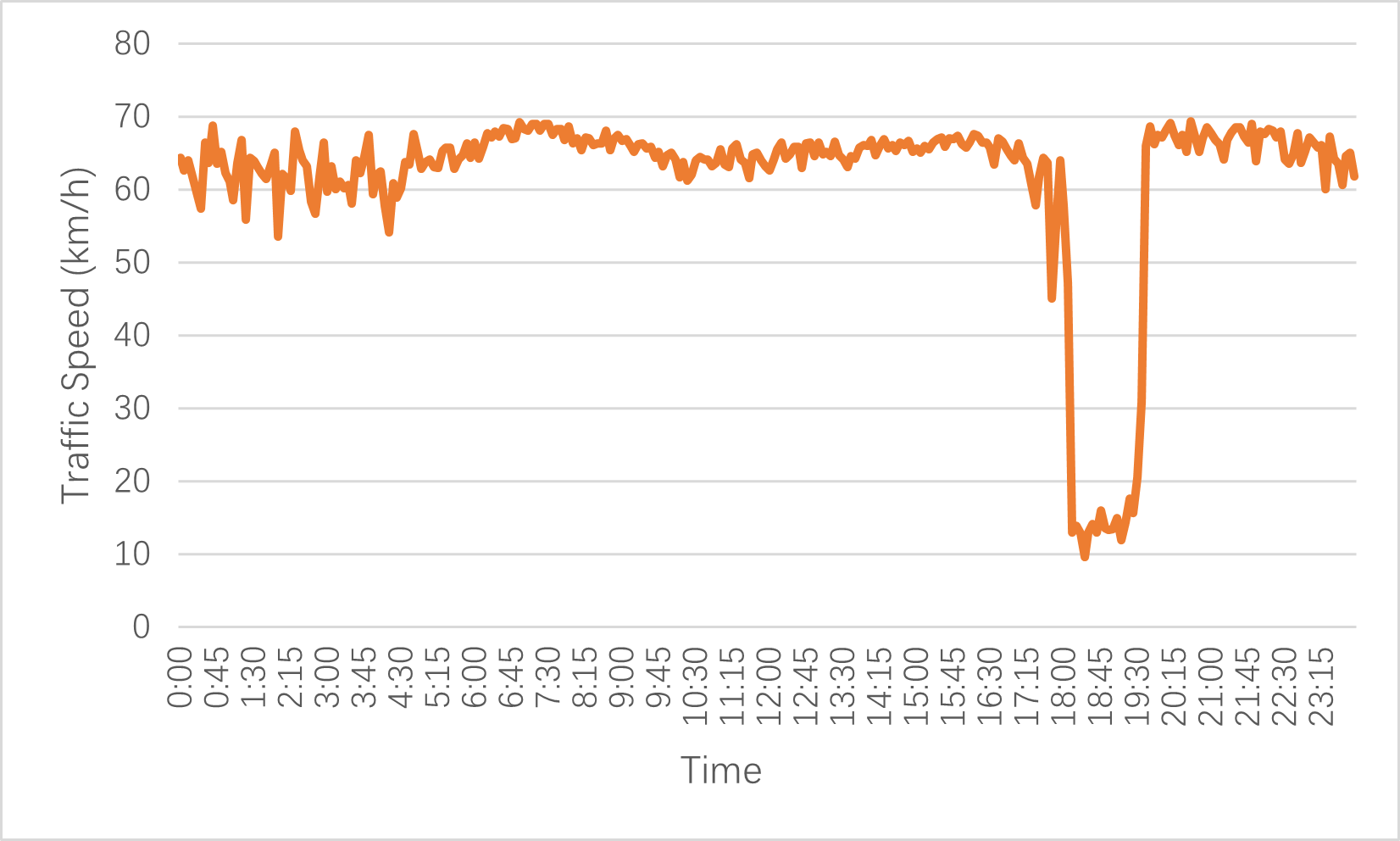}
	\caption{Traffic speed on Mar. 1st, Los Angeles}
	\label{fig51}
\end{figure}

For STGIN model, 80$\%$ of the data is used as the training set and the rest data is used as the testing set. The sequence samples are generated by sliding a window size of $T+T^{\prime}$, which the $T$ length is the input sequence and the $T^{\prime}$ length is the ground truth labeled. Since different horizons are selected for experiments, we manually set the input sequence to fit the requirement. The MSE loss function is selected as the loss function. We set attention heads to 4, batch size to 32, hidden dimensions to 32, training iteration to 500. For baseline models, we keep their settings unchanged as their papers reported.

\subsection{Evaluation Metrics}
There are three commonly used criteria applied here to evaluate the prediction effectiveness of STGIN model and baselines:
(1) Root Mean Squared Error (RMSE):
\begin{equation}
\text{RMSE} = \sqrt{\frac{1}{MN}\sum_{j=1}^M\sum_{i=1}^{N}(y_i^j-\hat{y_i^j})^2}
\label{rmse}
\end{equation}
RMSE is the standard deviation of the prediction errors.

(2) Mean Absolute Error (MAE):
\begin{equation}
\text{MAE} = \frac{1}{MN}\sum_{j=1}^M\sum_{i=1}^{N}| y_i^j-\hat{y_i^j} |
\label{mae}
\end{equation}
MAE is a measure of errors between ground truths and predictive value. The smaller the value of RMSE or MAE, the better predictive performance.

(3) Accuracy:

Accuracy is the most important index that identifies the prediction precision.
\begin{equation}
\text{Accuracy} = 1-\frac{{\|Y-\hat{Y}\|}_F}{{\|Y\|}_F}
\label{accuracy}
\end{equation}
where $y_i^j$ and $\hat{y_i^j}$ represent the ground truth traffic information and predicted value of the $j-th$ time instant in the $i-th$ location. M is the number of instants; N is the number of nodes; Y and $\hat{Y}$ represent the set of $y_i^j$ and $\hat{y_i^j}$ respectively.

\subsection{Baseline Settings}
Seven representative baselines containing RNN and Transformer are utilized for comparison to demonstrate the performance of the proposed model. The baselines are described as follows:

\begin{itemize}
	\item ARIMA: Auto-regressive Integrated Moving Average model is used to solve predictive problem by fitting time series data to estimate future data.
	
	\item SVR: Support Vector Regression model is a typical statistical regressive method used for traffic sequence regression by doing regression based on stable assumptions.
	
	\item GRU: Together with LSTM, they are two popular variants of RNN, which have been proven effective in traffic prediction problem and can alleviate the problem of gradient explosion and vanishing. GRU has a simpler structure than LSTM with faster training speed.
	
	\item STGCN: STGCN combines the GCN layer and the GRU layer while doing forecasting.
	
	\item Transformer: The classic Transformer model with the self-attention mechanism.
	
	\item GCN+Transformer: The combination of GCN and classic Transformer to extract spatial-temporal dependencies.
	
	\item Informer: A new Transformer variant proposed to process long sequence prediction issue without spatial dependencies extraction.
	
\end{itemize}

\subsection{Experiment Result and Analysis}

\begin{table*}[!htbp]
	\renewcommand\arraystretch{1.3}
	\centering
	\caption{Traffic Forecasting Results on SZ-taxi datasets}
	\resizebox{0.88\textwidth}{!}{%
		\begin{tabular}{|c|c|c|c|c|c|c|c|c|c|}
			\hline
			
			\multirow{2}*{Horizon} & \multirow{2}*{Criteria} &\multicolumn{8}{|c|}{Shenzhen Taxi Speed}\\
			\cline{3-10}
			{}&{}&{ARIMA}&{SVR}&{GRU}&{STGCN}&{Transformer}&{GCN+Transformer}&{Informer}&{STGIN}\\
			\hline
			
			\multirow{3}*{15min}&{MAE}&4.677&2.688&2.691&2.741&1.047& 2.848 &0.055&\textbf{0.050}\\
			
			&{RMSE}&6.790&4.163&4.094&3.966&3.117& 4.175 & 0.078 &\textbf{0.064}\\
			
			&{Accuracy}&0.385&0.710&0.718&0.727&0.768& 0.709 & 0.714 &\textbf{0.840}\\
			\hline

			\multirow{3}*{30min}&{MAE}&4.666&2.775&2.771&2.789&1.249& 2.828 & \textbf{0.055} &0.060\\
			
			&{RMSE}&6.771&4.216&4.075&4.005&3.884& 4.180 &0.079&\textbf{0.075}\\
			
			&{Accuracy}&0.385&0.706&0.720&0.724&0.756&0.709&0.712&\textbf{0.815}\\
			\hline

			\multirow{3}*{45min}&{MAE}&4.683&2.869&2.835&2.832&1.286&2.932&\textbf{0.055} &0.061\\
			
			&{RMSE}&6.835&4.363&4.127&3.062&4.361&4.266&0.079&\textbf{0.076}\\
			
			&{Accuracy}&0.381&0.699&0.700&0.705&0.730&0.703&0.711&\textbf{0.808}\\
			\hline
			
			\multirow{3}*{60min}&{MAE}&4.710&2.910&2.862&2.852&1.548&3.054&\textbf{0.056} &0.062\\
			
			&{RMSE}&6.999&4.486&4.396&4.139&4.369&4.319&0.080&\textbf{0.076}\\
			
			&{Accuracy}&0.379&0.652&0.678&0.692&0.729&0.699&0.709&\textbf{0.808}\\
			\hline
			
		\end{tabular}%
	}
	\label{table1}
\end{table*}

\begin{table*}[!htbp]
	\renewcommand\arraystretch{1.3}
	\centering
	\caption{Traffic Forecasting Results on Los-loop datasets}
	\resizebox{0.88\textwidth}{!}{%
		\begin{tabular}{|c|c|c|c|c|c|c|c|c|c|}
			\hline
			
			\multirow{2}*{Horizon} & \multirow{2}*{Criteria} &\multicolumn{8}{|c|}{Los Angeles Vehicle Speed}\\
			\cline{3-10}
			{}&{}&{ARIMA}&{SVR}&{GRU}&{STGCN}&{Transformer}&{GCN+Transformer}&{Informer}&{STGIN}\\
			\hline
			
			\multirow{3}*{15min}&{MAE}&7.689&3.725&3.651&3.747&1.844&3.039&0.099&\textbf{0.064}\\
			
			&{RMSE}&9.345&6.959&6.280&6.060&6.447&5.177&0.150&\textbf{0.098}\\
			
			&{Accuracy}&0.828&0.882&0.893&0.897&0.875&0.912&0.867&\textbf{0.920}\\
			\hline

			\multirow{3}*{30min}&{MAE}&7.695&4.504&4.519&4.602&2.114&3.525&0.103&\textbf{0.070}\\
			
			&{RMSE}&10.054&8.439&7.662&7.268&7.146&6.174&0.157&\textbf{0.106}\\
			
			&{Accuracy}&0.827&0.856&0.869&0.876&0.871&0.895&0.860&\textbf{0.912}\\
			\hline

			\multirow{3}*{45min}&{MAE}&7.736&5.285&5.168&5.162&2.409&4.080&0.106&\textbf{0.073}\\
			
			&{RMSE}&10.128&9.935&8.256&7.521&7.672&6.910&0.165&\textbf{0.115}\\
			
			&{Accuracy}&0.816&0.819&0.830&0.849&0.859&0.882&0.855&\textbf{0.903}\\
			\hline
			
			\multirow{3}*{60min}&{MAE}&7.803&5.719&5.619&5.592&2.520&4.507&0.110&\textbf{0.084}\\
			
			&{RMSE}&10.294&10.148&8.532&7.769&7.566&7.774&0.164&\textbf{0.125}\\
			
			&{Accuracy}&0.795&0.789&0.803&0.828&0.821&0.868&0.851&\textbf{0.893}\\
			\hline
			
		\end{tabular}%
	}
	\label{table2}
\end{table*}

Table 1 and table 2 show the results for SZ-Taxi and Los-loop datasets, respectively. For each dataset comparisons among baseline models and our proposed model, four different forecasting horizons of 15-min, 30-min, 45-min and 60-min are used to indicate the prediction performance. We can have the general judgment that STGIN model obtains the best performance under most metrics over 4 horizons and following features can be observed from the results:

\begin{itemize}
	\item It can be found deep neural network methods, from GRU to Informer, have better prediction performance than traditional regressive models, such as ARIMA and SVR, which further verifies the significance of temporal dependencies extraction. It is difficult for regressive methods to do the forecasting based on nonstationary and comprehensive time-series data. For instance, in 15-min forecasting task of Los-loop dataset, STGIN achieves 9.7$\%$ prediction accuracy improvement than that of ARIMA and RMSE error of STGIN is reduced by 49.7$\%$ compared with the ARIMA model.
	
	\item We can observe STGIN achieves the best prediction accuracy over four horizons and the performance is stable as the extension of testing length. The longest window even reaches 60-min, which has great reference value in the real traffic management. From the results, we can know our proposed method can be utilized for both short-term and long-term prediction reliably.
	
	\item To show the significance of spatial dependencies extraction when doing the forecasting, our experiments set the internal comparison for each main temporal extraction model. For two datasets, We can observe obvious prediction accuracy improvements of STGIN than Informer, GCN+Transformer than Transformer and STGCN than GRU over all four horizons for both datasets, indicating that GNN layer can help to capture spatial dependencies from sequential data and achieve better understanding of the hidden pattern.
	
\end{itemize}

\subsection{STGIN Model Interpretation}
A case study here is carried out to show how STGIN model works on long sequence traffic prediction. We select one node in SZ-taxi dataset. One short and one long horizons (15-min and 60-min) ahead prediction results are visualized with their ground truths. And Fig.11 - Fig.14 show the 15-min prediction results improvement process as the training epochs increases. Fig.15 - Fig.18 show the 60-min prediction results improvement process.

\begin{figure}[H]
	\centering
	\includegraphics[width=0.8\linewidth]{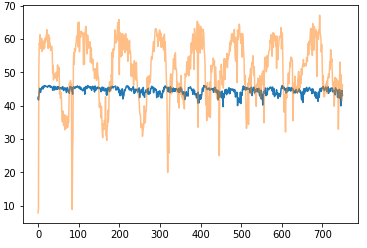}
	\caption{2-Epoch 15-min training results on SZ-taxi}
	\label{fig11}
\end{figure}

\begin{figure}[H]
	\centering
	\includegraphics[width=0.8\linewidth]{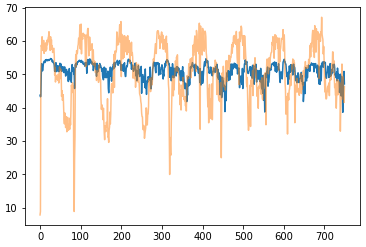}
	\caption{4-Epoch 15-min training results on SZ-taxi}
	\label{fig12}
\end{figure}

\begin{figure}[H]
	\centering
	\includegraphics[width=0.8\linewidth]{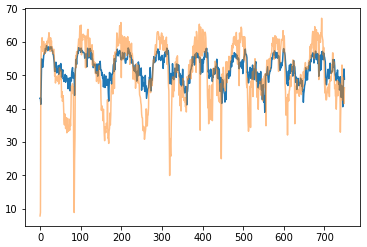}
	\caption{8-Epoch 15-min training results on SZ-taxi}
	\label{fig13}
\end{figure}

\begin{figure}[H]
	\centering
	\includegraphics[width=0.8\linewidth]{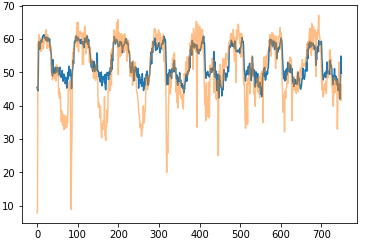}
	\caption{16-Epoch 15-min training results on SZ-taxi}
	\label{fig14}
\end{figure}

\begin{figure}[H]
	\centering
	\includegraphics[width=0.8\linewidth]{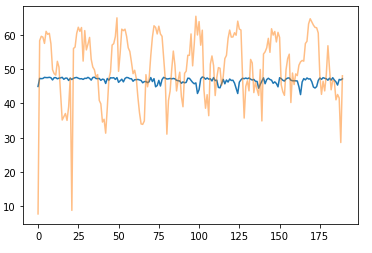}
	\caption{10-Epoch 60-min training results on SZ-taxi}
	\label{fig15}
\end{figure}

\begin{figure}[H]
	\centering
	\includegraphics[width=0.8\linewidth]{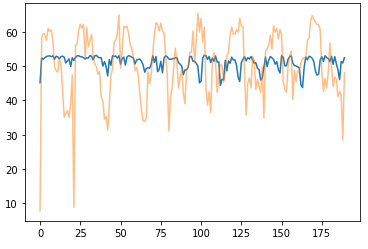}
	\caption{20-Epoch 60-min training results on SZ-taxi}
	\label{fig16}
\end{figure}

\begin{figure}[H]
	\centering
	\includegraphics[width=0.8\linewidth]{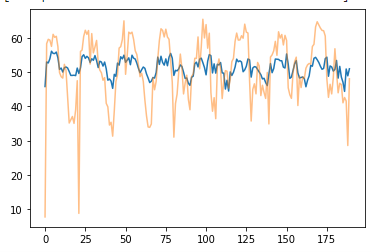}
	\caption{40-Epoch 60-min training results on SZ-taxi}
	\label{fig17}
\end{figure}

\begin{figure}[H]
	\centering
	\includegraphics[width=0.8\linewidth]{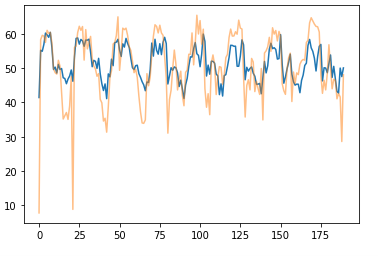}
	\caption{80-Epoch 60-min training results on SZ-taxi}
	\label{fig18}
\end{figure}

For the 15-min ahead case, it can be observed that the prediction performance is not good at the early stage (before 2 epochs) in Fig.7. Most data changing trends are not captured and only the central values of the fluctuation range are reflected. However, STGIN perceives the periodicity and data shift rapidly, only 2 epochs later, the model can catch the changing curve, even though some abrupt changes cannot be fitted well. As more training epochs are executed, the prediction performance are improved continuously and STGIN can fit the ground truth quite well after 16 epochs.

For the 60-min ahead case, which is quite a challenge in real-world applications. However, we can observe STGIN still can learn and capture the feature quickly. The results further prove the long-term prediction capability of the proposed model.

\section{Conclusion}
The study proposes a novel spatial-temporal Transformer-based framework called STGIN to address the challenge of long sequence traffic forecasting. We integrate the GAT and Informer layers to capture traffic information's spatial and temporal relationships. Through the use of the attention mechanism, the Informer layer is able to acquire global temporal information when processing long sequence time series data. Comparing STGIN against other extensively used baseline models such as GRU, Transformer, and other validated frameworks, we demonstrate its efficacy with two real-word datasets. STGIN delivers the best prediction performance over a range of horizons and datasets while maintaining a high degree of stability. Additionally, the capacity for long-term forecasting has been shown. In summary, the STGIN model successfully completed long-term accurate traffic forecasting by collecting spatial-temporal characteristics, indicating that it has a great promise for future real-world traffic forecasting tasks.

\section*{Acknowledgment}
This study is supported by the RIE2020 Industry Alignment Fund – Industry Collaboration Projects (IAF-ICP) Funding Initiative, as well as cash and in-kind contribution from the industry partner(s) and A*STAR by its RIE2020 Advanced Manufacturing and Engineering (AME) Industry Alignment Fund C Pre-Positioning (IAF-PP) (Award A19D6a0053).

\ifCLASSOPTIONcaptionsoff
  \newpage
\fi


\singlespacing
\bibliographystyle{IEEEtran}
\bibliography{autosam}


\begin{IEEEbiography}[{\includegraphics[width=1in,height=1.25in,clip,keepaspectratio]{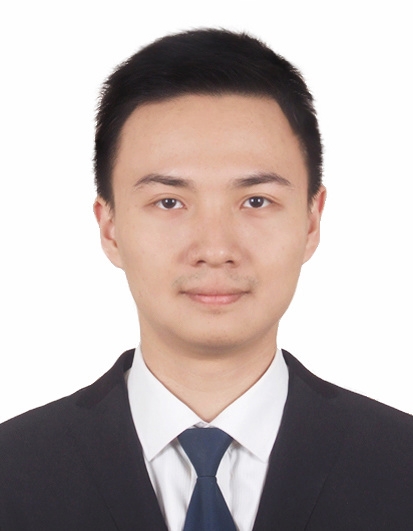}}]{Ruikang Luo}
	received the B.E. degree from the School of Electrical and Electronic Engineering, Nanyang Technological University, Singapore. He is currently currently pursuing the Ph.D. degree with the School of Electrical and Electronic Engineering, Nanyang Technological University, Singapore. His research interests include long-term traffic forecasting based on spatiotemporal data and artificial intelligence.
\end{IEEEbiography}

\begin{IEEEbiography}[{\includegraphics[width=1in,height=1.25in,clip,keepaspectratio]{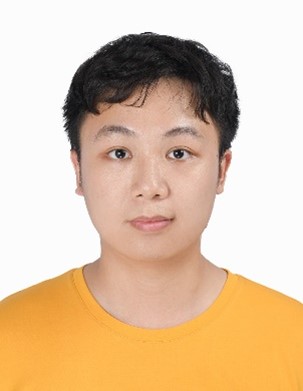}}]{Yaofeng Song}
	received the bachelor degree from the school of Automation Science and Engineering in South China University of Technology. Currently he is a Msc student in the school of Electrical and Electronic Engineering in Nanyang Technological University, Singapore. His research interests invlove deep learning based traffic forecasting.
\end{IEEEbiography}

\begin{IEEEbiography}[{\includegraphics[width=1in,height=1.25in,clip,keepaspectratio]{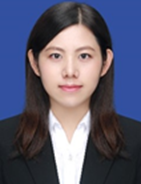}}]{Liping Huang}
	Huang Liping obtained her Ph. D, and Master of Computer Science from Jilin University in 2018 and 2014, respectively. She has been working as a research fellow at Nanyang Technological University since 2019 June. Dr. Huang’s research interests include spatial and temporal data mining, mobility data pattern recognition, time series prediction, machine learning, and job shop scheduling. In the aforementioned areas, she has more than twenty publications and serves as the reviewer of multiple journals, such as IEEE T-Big Data, IEEE T-ETCI, et al.
\end{IEEEbiography}

\begin{IEEEbiography}[{\includegraphics[width=1in,height=1.25in,clip,keepaspectratio]{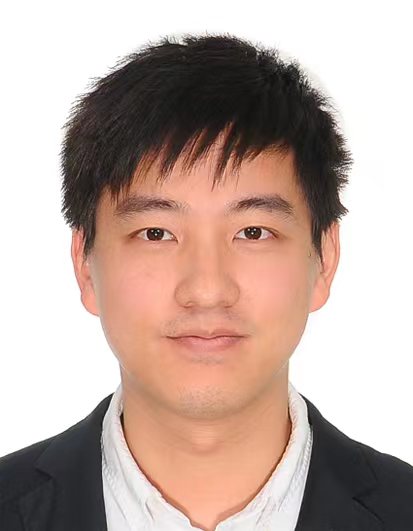}}]{Yicheng Zhang}
	Yicheng Zhang received the Bachelor of Engineering in Automation from Hefei University of Technology in 2011, the Master of Engineering degree in Pattern Recognition and Intelligent Systems from University of Science and Technology of China in 2014, and the PhD degree in Electrical and Electronic Engineering from Nanyang Technological University, Singapore in 2019. He is currently a research scientist at the Institute for Infocomm Research (I2R) in the Agency for Science, Technology and Research, Singapore (A*STAR). Before joining I2R, he was a research associate affiliated with Rolls-Royce @ NTU Corp Lab. He has participated in many industrial and research projects funded by National Research Foundation Singapore, A*STAR, Land Transport Authority, and Civil Aviation Authority of Singapore. He published more than 70 research papers in journals and peer-reviewed conferences. He received the IEEE Intelligent Transportation Systems Society (ITSS) Young Professionals Traveling Scholarship in 2019 during IEEE ITSC, and as a team member, received Singapore Public Sector Transformation Award in 2020.
\end{IEEEbiography}

\begin{IEEEbiography}[{\includegraphics[width=1in,height=1.25in,clip,keepaspectratio]{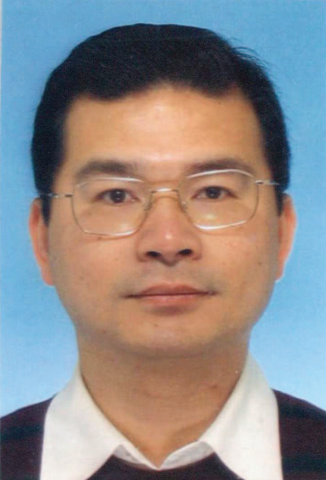}}]{Rong Su}
	received the M.A.Sc. and Ph.D. degrees both	in electrical engineering from the University of Toronto, Toronto, Canada, in 2000 and 2004 respectively.	He is affiliated with the School of Electrical and Electronic Engineering, Nanyang Technological University, Singapore. His research interests include modeling, fault diagnosis and supervisory control of discrete-event dynamic systems. Dr. Su has been a member of IFAC technical committee on discrete event and hybrid systems (TC 1.3) since 2005.
\end{IEEEbiography}

\end{document}